\documentclass{llncs}
\usepackage{llncsdoc}
\usepackage{amsmath}
\usepackage{graphicx}

\begin{document}
\title{Advances in Feature Selection with Mutual Information}
\titlerunning{Advances in Feature Selection with Mutual Information}  
%
\author{Michel Verleysen\inst{1} \and Fabrice Rossi\inst{2} \and Damien Fran\c{c}ois\inst{1}}
\authorrunning{Verleysen, Rossi, and  Fran\c{c}ois}   
%
%
\institute{Universit\'e catholique de Louvain, Machine Learning Group,\and
INRIA Rocquencourt, Domaine de Voluceau, Rocquencourt, B.P. 105,
78153 Le Chesnay Cedex, France,\\
\email{Michel.Verleysen@uclouvain.be, Fabrice.Rossi@inria.fr, Damien.Francois@uclouvain.be}}

\maketitle              

\begin{abstract}
The selection of features that are relevant for a prediction or classification problem is an important problem in many domains involving high-dimensional data.  Selecting features helps fighting the curse of dimensionality, improving the performances of prediction or classification methods, and interpreting the application.  In a nonlinear context, the mutual information is widely used as relevance criterion for features and sets of features.  Nevertheless, it suffers from at least three major limitations: mutual information estimators depend on smoothing parameters, there is no theoretically justified stopping criterion in the feature selection greedy procedure, and the estimation itself suffers from the curse of dimensionality.  This chapter shows how to deal with these problems.  The two first ones are addressed by using resampling techniques that provide a statistical basis to select the estimator parameters and to stop the search procedure.  The third one is addressed by modifying the mutual information criterion into a measure of how features are complementary (and not only informative) for the problem at hand.
\end{abstract}
\section{Introduction}

High-dimensional data are nowadays found in many applications areas: image and
signal processing, chemometrics, biological and medical data analysis, and many
others.  The availability of low cost sensors and other ways to measure
information, and the increased capacity and lower cost of storage equipments,
facilitate the simultaneous measurement of many features, the idea being that
adding features can only increase the information at disposal for further
analysis. 

The problem is that high-dimensional data are in general more difficult to
analyse.  Standard data analysis tools either fail when applied to
high-dimensional data, or provide meaningless results.  Difficulties related
to handling high-dimensional data are usually gathered under the \emph{curse
  of dimensionality} terms, which gather many phenomena usually having
counter-intuitive mathematical or geometrical interpretation.  The curse of
dimensionality already concerns simple phenomena, like colinearity.  In many
real-world high-dimensional problems, some features are highly correlated.
But if the number of features exceeds the number of measured data, even a
simple linear model will lead to an undetermined problem (more parameters to
fit than equations).  Other difficulties related to the curse of
dimensionality arise  in more common situations, when the dimension of the
data space is high even if many data are available for fitting or learning.
For example, data analysis tools which use Euclidean distances between data or
representatives, or any kind of Minkowski or fractional distance (i.e. most
tools) suffer from the fact that distances concentrate in high-dimensional spaces
(distances between two random close points and between two random far ones
tend to converge to the same value, in average). 

Facing these difficulties, data analysis tools must address two ways to
counteract them.  One is to develop tools that are able to model
high-dimensional data with a number of (effective) parameters which is lower
than the dimension of the space.  As an example, Support-Vector Machines enter
into this category.  The other way is to decrease in some way the dimension of
the data space, without significant loss of information.  The two ways are
complementary, as the first one addresses the algorithms while the second
preprocesses the data themselves.  Two possibilities also exist to reduce the
dimensionality of the data space: features (dimensions) can be selected, or
combined.  Feature combination means to \emph{project} data, either linearly
(Principal Component Analysis, Linear Discriminant Analysis, etc.) or
nonlinearly.  Selecting features, i.e. keeping some of the original features
as such, and discarding others, is a priori less powerful than projection (it
is a particular case).  However, it has a number of advantages, mainly when
interpretation is sought.  Indeed after selection the resulting features are
among the original ones, which allows the data analyst to interact with the
application provider.  For example, discarding features may help avoiding to
collect useless (possibly costly) features in a further measument campaign.
Obtaining relevances for the original features may also help the application
specialist to interpret the data analysis results, etc.  Another reason to
prefer selection to projection in some circumstances, is when the dimension
of the data is really high, and the relations between features known or
identified to be strongly nonlinear.  In this case indeed linear projection
tools cannot be used; and while nonlinear dimensionality reduction is nowadays
widely used for data visualization, its use in quantitative data preprocessing
remains limited because of the lack of commonly accepted standard method,
the need for expertise to use most existing tools and the computational cost
of some of the methods.

This chapter deals with feature selection, based on mutual information between
features.    The following of this chapter is organized as follows. Section \ref{sec:ingredients} introduces the problem of feature selection and the main ingredients of a selection procedure.  Section \ref{sec:MI} details the Mutual Information relevance criterion, and the difficulties related to its estimation.  Section \ref{sec:solutions} shows how to solve these issues, in particular how to choose the smoothing parameter in the Mutual Information estimator, how to stop the greedy search procedure, and how to extend the mùutual information concept by using nearest neighbor ranks when the dimension of the search space increases.

\section{The two ingredients of feature selection}
\label{sec:ingredients}
Feature selection aims at reducing the dimensionality of data.  It consists in selecting \emph{relevant} variables (or features) among the set of original ones.  The relevance has to be measured in an objective way, through an appropriate and well-defined criterion.  However, defining a criterion does not solve the feature selection problem.  As the number of initial features is usually large, it is computationally impossible to test all possible subsets of them, even if the criterion to measure the relevance is simple to evaluate.  In addition to the definition of the criterion, there is thus a need to define a search procedure among all possible subsets.  The relevance criterion and the greedy search procedure are the two basic ingredients of feature selection.

Note that in some situations, feature selection does not aim only at selecting
features among the original ones.  In some cases indeed, potentially relevant
features are not known in advance, and must be extracted or created from the
raw data.  Think for example of data being curves, as in spectroscopy, in
hysteresis curve analysis, or more generally in the processing of functions.
In this case the dimension of the data is infinite, and a first choice must
consist in extracting a finite number of original features.  Curve sampling
may be an answer to this question, but other features, as integrals, area
under curve, derivatives, etc. may give appropriate information for the
problem too. It may thus reveal interesting to first extract a large number of
features in a more or less blind way from the original data, and then to use
feature selection to select those that are most relevant, in an objective
way. 

In addition to choosing a relevance criterion and a greedy procedure, a number
of other issues have to be addressed.  First, one has to define on which
features to apply the criterion.  For example, if the criterion is
correlation, is it better to keep features that are highly correlated to the
output (and to drop the other ones), or to drop features that are highly
correlated between them (and to keep uncorrelated ones)?  Both ideas are
reasonable, and will lead to different selections. 

Another key issue is simply whether to use a criterion or not.  If the goal of
feature selection is to use the reduced feature set as input to a prediction
or classification model, why not to use the model itself as a criterion?  In
other words, why not fitting a model on each possible subset (resulting from
the greedy search), instead of using a criterion that will probably result in
measuring relevance in a different way as the model would do?  Using the
model is usually referred to as a \emph{wrapper} approach, while using an
alternative criterion is a \emph{filter} approach.  In theory, there is
nothing better than using the model itself, as the final goal is model
performances.  However, the wrapper way may have two drawbacks: first it could
be computationally too intensive for example when using nonlinear neural
networks or machine learning tools that require tedious learning.  Secondly,
when the stochastic nature of the tools makes that their results vary
according to initial conditions or other parameters, the results may not be
unique, which results in a noisy estimation of the relevance and the need for
further simulations to reduce this noise.  The main goal of criteria in
filters is then to \emph{facilitate} the measure of feature relevance,
rather than to provide a unique and unquestionable way of evaluation.  This
must be kept in mind when designing both the relevance criterion and the greedy
procedure: both will act as compromises between adequateness (with the final
goal of model performances) and the computational complexity.

Most of the above issues are extensively discussed in the large scientific
literature about feature selection.  One issue which is much less discussed is
how to evaluate the criterion.  An efficient criterion must measure any kind
of relation between features, not only linear relations.  Such a nonlinear
criterion is however a (simplified) data model by itself, and requires to fix
some design parameters.  How to fix these parameters is an important question
too, as an inappropriate choice may lead to wrong relevance estimations.

This chapter mainly deals with the last question, i.e. how to estimate in
practice, and efficiently, the relevance criterion.  Choices that are made
concerning the criterion itself and the greedy procedure are as follows.

As the relevance criterion must be able to evaluate any relation between
features, and not only linear relations, the correlation is not appropriate.
A nonlinear extension to correlation, borrowed from the information theory, is
the mutual information (MI).  The mathematical definition of MI and its
estimation will be detailed in the next section. 

Feature selection necessitates to select \emph{sets of} features.  This means
that it is the relevance of the sets that must be evaluated, rather than the
relevance of the features in the set.  Indeed evaluating individually the
relevance of single features would result in similar relevances; if two highly
correlated, but highly relevant too, features are contained in the original
set, they will then both be selected, while selecting one would have been
sufficient for the prediction or classification model.  Evaluating sets of
features means in other words, to be able to evaluate the relevance of a
multi-dimensional variable (a vector), instead of a scalar one only.  Again MI
is appropriate with this respect, as detailed in the next section.

Finally, many greedy procedures are proposed in the literature.  While several
variants exist, they can be roughly categorized in \emph{forward} and
\emph{backward} procedures; the former means that the set is built from
scratch by adding relevant features at each step, while the latter proceeds by
using the whole set of initial features and removing irrelevant ones.  Both
solutions have their respective advantages and drawbacks.  A drawback of the
forward procedure is that the initial choices (when few features are
concerned) influence the final choice, and may reveal suboptimal.  However,
the forward procedure has an important advantage: the maximum size of the
vectors (feature sets) that have to be evaluated by the criterion is equal to
the final set size.  In the backward approach, the maximum size is the one in
the initial step, i.e. the size of the initial feature set.  As it will be
seen in the next section, the evaluation of the criterion is also made more
difficult because of the curse of dimensionality; working in smaller space
dimensions is thus preferred, what justifies the choice for a forward
approach.

\section{Feature selection with Mutual Information}
\label{sec:MI}

A prediction (or classification) model aims to reduce the uncertainty on the
output, the dependent variable.  As mentioned in the previous section, a good
criterion to evaluate the relevance of a (set of) feature(s) is nothing else
than a simplified prediction model. A natural idea is then to measure the
uncertainty of the output, given the fact that the inputs (independent
variables) are known.  The formalism below is inspired from
\cite{RossiCILS06}.

\subsection{Mutual information definition}

A powerful formalization of the uncertainty of a random variable is Shannon's
entropy.  Let $X$ and $Y$ be two random variables; both may be multidimensional
(vectors). Let $\mu_X(x)$ and $\mu_Y(y)$ the (marginal) probability density
functions (pdf) of $X$ and $Y$, respectively, and $\mu_{X,Y}(x,y)$ the joint
pdf of the $(X,Y)$ variable.  The entropies of $X$ and of $Y$, which measures
the uncertainty on these variables, are defined respectively as
\begin{eqnarray}
  H(X)&=&-\int \mu_X(x) \log{\mu_X(x)} dx,\\
  H(Y)&=&-\int \mu_Y(y) \log{\mu_Y(y)} dy.
\end{eqnarray}
If $Y$ depends on $X$, the uncertainty on $Y$ is reduced when $X$ is known.
This is formalized through the concept of conditional entropy: 
\begin{equation}
  H(Y|X)=-\int \mu_X(x) \int \mu_Y(y|X=x) \log{\mu_Y(y|X=x)} dydx.
\end{equation}
The Mutual Information (MI) then measures the reduction in the uncertainty on $Y$ resulting from the knowledge of $X$:
\begin{equation}
  MI(X,Y)=H(Y)-H(Y|X).
\end{equation}
It can easily be verified that the MI is symmetric:
\begin{equation}
  MI(X,Y)=MI(Y,X)=H(Y)-H(Y|X)=H(X)-H(X|Y);
\end{equation}
it can be computed from the entropies:
\begin{equation}
  MI(X,Y)=H(X)+H(Y)-H(X,Y),
  \label{eq:MIfromH}
\end{equation}
and is equal to the Kullback-Leibler divergence between the joint pdf and the product of the marginal pdfs:
\begin{equation}
  I(X,Y)=\int{\int{\mu_{X,Y}(x,y)\log{\frac{\mu_{X,Y}(x,y)}{\mu_X(x)\mu_Y(y)}}}}.
    \label{eq:MIbyKL}
\end{equation}
In theory as the pdfs $\mu_X(x)$ and $\mu_Y(y)$ may be computed from the joint
one $\mu_{X,Y}(x,y)$ (by integrating over the second variable), one only needs
$\mu_{X,Y}(x,y)$ in order to compute the MI between $X$ and $Y$.

\subsection{Mutual Information estimation}
According to the above equations, the estimation of the MI between $X$ and $Y$
may be carried out in a number of ways.  For instance, equation \eqref{eq:MIfromH} may be used after the entropies of $X$, $Y$ and $X,Y$ are estimated, or the
Kullback-Leibler divergence between the pdfs may be used as in
equation \eqref{eq:MIbyKL}. 

The latter solution necessitates to estimate the pdfs from the know sample
(the measured data).  Many methods exist to estimate pdfs, including
histograms and kernel-based approximations (Parzen windows), see
e.g. \cite{Scott:1992}.  However, these approaches are inherently restricted
to low-dimensional variables.  If the dimension of $X$ exceeds let's say
three, histograms and kernel-based pdf estimation requires a prohibitive
number of data; this is a direct consequence of the curse of dimensionality
and the so-called empty space phenomenon.  However, as mentioned in the
previous section, the MI will have to be estimated on sets of features (of
increasing dimension in the case of a forward procedure). Histograms and
kernel-based approximators become rapidly inappropriate for this reason.

Although not all problems related to the curse of dimensionality are solved in
this way, it appears that directly estimating the entropies is a better
solution, at least if an efficient estimator is used.  Intuitively, the
uncertainty on a variable is high when the distribution is flat and small when
it has high peaks.  A distribution with peaks means that neighbors (or
successive values in the case of a scalar variable) are very close, while in
a flat distribution the distance between a point and its neighbors is larger.
Of course this intuitive concept only applies if there is a finite number of
samples; this is precisely the situation where it is needed to estimate the
entropy rather than using its integral definition.  This idea is formalized in
the Kozachenko-Leonenko estimator for differential Shannon entropy
\cite{Koza}:
\begin{equation}
  \hat{H}(X)=-\psi(K)+\psi(N)+\log{c_D}+\frac{D}{N} \sum_{n=1}^N {\log{\epsilon(n,K)}}
  \label{entropy_estimator}
\end{equation}  
where $N$ is the number of samples $x_n$ in the data set, $D$ is the
dimensionality of $X$, $c_D$ is the volume of a unitary ball in a
$D$-dimensional space, and $\epsilon(n,K)$ is twice the distance from $x_n$ to
its $K$-th neighbour.  $K$ is a parameter of the estimator, and $\psi$ is the
digamma function given by
\begin{equation}
  \psi(t)=\frac{\Gamma'(t)}{\Gamma(t)}=\frac{d}{dt}\ln\Gamma(t),
\end{equation}
with
\begin{equation}
  \Gamma(t)=\int_0^{\infty}{u^{t-1}e^{-u}du}.
\end{equation}
The same intuitive idea of $K$-th nearest neighbor is at the basis of an
estimator of the MI between $X$ and $Y$.  The MI is aimed to measure the loss
of uncertainty on $Y$ when $X$ is known.  In other words, this means to
answer the question whether some (approximate) knowledge on the value of $X$
may help identifying what can be the possible values for $Y$.  This is only
feasible if there exists a certain notion of continuity or smoothness when
looking to $Y$ with respect to $X$. Therefore, close values in $X$ should
result in corresponding close values in $Y$.  This is again a matter of
$K$-nearest neighbors: for a specific data point, if its neighbors in the $X$
and $Y$ spaces correspond to the same data, then knowing $X$ helps in knowing
$Y$, which reflects a high MI.

More formally, let us define the joint variable $Z=(X,Y)$, and $z_n=(x_n,y_n),
1\leq n\leq N$ the available data. Next, we define the norm in the $Z$ space
as the maximum norm between the $X$ and $Y$ components; if $z_n=(x_n,y_n)$ and
$z_m=(x_m,y_m)$, then
\begin{equation}
  \left\|z_n-z_m\right\|_{\infty}=\max(\left\|x_n-x_m\right\|,\left\|y_n-y_m\right\|),
\end{equation}
where the norms in the $X$ and $Y$ spaces are the natural ones. Then
$z_{K(n)}$ is defined as the $K$-nearest neighbor of $z_n$ (measured in the
$Z$ space).  $z_{K(n)}$ can be decomposed in its $x$ and $y$ parts as
$z_{K(n)}=\left(x_{K(n)},y_{K(n)}\right)$; note however that $x_{K(n)}$ and
$y_{K(n)}$ are not (necessarily) the $K$-nearest neighbors of $x_n$ and $y_n$
respectively, with $z_n=\left(x_n,y_n\right)$.

Finally, we denote
\begin{equation}
  \epsilon_n=\left\|z_n-z_{K(n)}\right\|_{\infty}
\end{equation}
the distance between $z_n$ and its $K$-nearest neighbor.  We can now count the
number $\tau_x(n)$ of points in $X$ whose distance from $x_n$ is strictly less
than $\epsilon_n$, and similarly the number $\tau_y(n)$ of points in $Y$ whose
distance from $y_n$ is strictly less than $\epsilon_n$.  It can then be shown
\cite{Kraskov:2004} that $MI(X,Y)$ can be estimated as:
\begin{equation}
  \widehat{MI}(X,Y)=\psi(K)+\psi(N)-\frac{1}{N} \sum_{n=1}^N {\left[\psi(\tau_x(n))+\psi(\tau_y(n))\right]}.
  \label{MI_estimator}
\end{equation} 
As with the Kozachenko-Leonenko estimator for differential entropy, $K$ is a
parameter of the algorithm and must be set with care to obtain an acceptable
MI estimation.  With a small value of $K$, the estimator has a small bias but
a high variance, while a large value of $K$ leads to a small variance but a
high bias.

In summary, while the estimator \eqref{MI_estimator} may be efficiently used
to measure the mutual information between $X$ and $Y$ (therefore the relevance
of $X$ to predict $Y$), it still suffers from two limitations.  Firstly, there
is a parameter ($K$) in the estimator that must be chosen with care.
Secondly, it is anticipated that the accuracy of the estimator will decrease
when the dimension of the $X$ space increases, i.e. along the steps of the
forward procedure.  These two limitations will be addressed further in this
contribution.

\subsection{Greedy selection procedure}

Suppose that $M$ features are initially available.  As already mentioned in
Section \ref{sec:ingredients}, even if the relevance criterion was
well-defined and easy to estimate, it is usually not possible to test all
$2^M-1$ non-empty subsets of features in order to select the best one.  There
is thus a need for a greedy procedure to reduce the search space, the aim
being to have a good compromise between the computation time (or the number of
tested subsets) and the potential usefulness of the considered subsets.  In
addition, the last limitation mentioned in the previous subsection gives the
preference to greedy search avoiding subsets with a too large number of
features.

With these goals in mind, it is suggested to use a simple forward procedure.
The use of a backward procedure (starting from the whole set of $M$ features)
is not considered to avoid having to evaluate mutual information on
$M$-dimensional vectors.

The forward search consists first in  selecting the feature that maximizes the mutual information with the output $Y$:
\begin{equation}
\label{eq:first_selected_MI}
X_{s_1}=\arg\max_{X_j,\, 1\leq j\leq M}\widehat{MI}\left(X_j,Y\right).
\end{equation}
Then in step $t$ ($t\geq 2$), the $t$-th features is selected as
\begin{equation}
\label{eq:next_selected_MI}
X_{s_t}=\arg\max_{X_j,\, 1\leq j\leq M,\, j\notin \{s_1,s_2,\ldots,s_{t-1}\}}\widehat{MI}\left(\left\{X_{s_1},X_{s_2},\ldots,X_{s_{t-1}}, X_j\right\},Y\right).
\end{equation}

Selecting features incrementally as defined by equations
\eqref{eq:first_selected_MI} and \eqref{eq:next_selected_MI} makes the
assumption that once a feature is selected, it should remain in the final set.
Obviously, this can lead to a suboptimal solution: it is not because the first
feature (for example) is selected according to equation
\eqref{eq:first_selected_MI} that the optimal subset necessarily contains this
feature.  In other words, the selection process may be stuck in a local
minimum.  One way to decrease the probability of being stuck in a local
minimum is to consider the removing of a single feature at each step of the
algorithm.  Indeed, there is no reason that a selected feature (for example
the first one according to equation \eqref{eq:first_selected_MI}) belongs to
the optimal subset.  Giving the possibility to remove a feature that has
become useless after some step of the procedure is thus advantageous, while
the increased computational cost is low.  More formally, the feature defined
as
\begin{equation}
\label{eq:removed_MI}
X_{s_d}=\arg\max_{X_j,\ 1\leq j\leq t-1}\widehat{MI}\left(\left\{X_{s_1},X_{s_2},\ldots,X_{s_{j-1}}, X_{s_{j+1}},\ldots,X_{s_t}\right\},Y\right)
\end{equation}
is removed if
\begin{equation}
\label{eq:removed_condition}
\widehat{MI}\left(\left\{X_{s_1},\ldots,X_{s_{d-1}}, X_{s_{d+1}},\ldots,X_{s_t}\right\},Y\right) > \widehat{MI}\left(\left\{X_{s_1},\ldots,X_{s_t}\right\},Y\right).
\end{equation}
Of course, the idea or removing features if the removal leads to an increased
MI may be extended to several features at each step.  However, this is nothing
else than extending the search space of subsets.  The forward-backward
procedure consisting in considering the removal of only one feature at each
step is thus a good compromise between expected performances and computational
cost.

Though the above suggestion seems to be appealing, and is used in many
state-of-the-art works, it is theoretically not sound.  Indeed, it can easily
be shown that the mutual information can only increase if a supplementary
variable is added to a set \cite{Cover:1991}.  The fact that equation
\eqref{eq:removed_condition} may hold in practice is only due to the fact that
equations \eqref{eq:removed_MI} and \eqref{eq:removed_condition} involve
estimations of the MI, and not the theoretical values.  The question is then
whether it is legitimate to think that condition \eqref{eq:removed_condition}
will effectively lead to the removal of unnecessary features, or if this
condition will be fulfilled by chance, without a sound link to the non-relevance
of the removed features.

Even if the backward procedure is not used, the same problem appears.  In
theory indeed, if equation \eqref{eq:next_selected_MI} is applied repeatedly
with the true MI instead of the estimated one, the MI will increase at each
step.  There is thus no stopping criterion, and without additional constraint
the procedure will result in the full set of $M$ initial features!  The
traditional way is then to stop when the estimation of the MI begins to
decrease.  This leads to the same question whether the decrease of the
estimated value is only due to a bias or noise in the estimator, or has a
sound link to the non or low relevance of a feature.

\subsection{The problems to solve}

To conclude this section, coupling the use of an estimator of the mutual
information, even if this estimator is efficient, to a greedy procedure raises
several questions and problems.  First, the estimator includes (as any
estimator) a smoothing parameter that has to be set with care.  Secondly, the
dimension of the vectors from which a MI has to be estimated may have an
influence on the quality of the estimation.  Finally, the greedy procedure
(forward, or forward-backward) needs a stopping criterion.  In the following
section, we propose to solve all these issues together by the adequate use of
resampling methods.  We also introduce an improvement to the concept of mutual
information, when used to measure the relevance of a (set of) features.

\section{Improving the feature selection by MI}
\label{sec:solutions}

In this section, we first address the problem of setting the smoothing
parameter in the MI estimator, by using resampling methods.  Secondly, we show
how using the same resampling method provides a natural and sound stopping
criterion for the greedy procedure.  Finally, we show how to improve the
concept of MI, by introducing a conditional redundancy concept. 

\subsection{Parameter setting in the MI estimation}\label{subsection:Kchoice}

The estimator defined by equation \eqref{MI_estimator} faces a classical
bias/variance dilemma. While the estimator is known to be consistent (see
\cite{GoriaEtAlJNPS2005}), it is only asymptotically unbiased and can
therefore be biased on a finite sample. Moreover, as observed in
\cite{Kraskov:2004}, the number of neighbors $K$ acts as a smoothing
parameter for the estimator: a small value of $K$ leads to a large variance
and a small bias, while a large value of $K$ has the opposite effects (large
bias and small variance).

Choosing $K$ consists therefore in balancing the two sources of inaccuracy in
the estimator. Both problems are addressed by resampling techniques. A
cross-validation approach is used to evaluate the variance of the estimator
while a permutation method provide some baseline value of the mutual
information that can reduce the influence of the bias. Then $K$ is chosen so
as to maximize the significance of the high MI values produced by the
estimator. 

The first step of this solution consists in evaluating the variance of the
estimator. This is done by producing ``new'' datasets drawn from the original
dataset $\Omega =\{(x_n,y_n)\}, 1\leq n\leq N$. As the chosen estimator
strongly overestimates the mutual information when submitted replicated
observations, the subsets cannot be obtained via random sampling with
replacement (i.e. bootstrap samples), but are on the contrary strict subsets
of $\Omega$. We use a cross-validation strategy: $\Omega$ is split randomly
into $S$ non-overlapping subsets $U_1,\ldots,U_S$ of approximately equal sizes
that form a partition of $\Omega$. Then $S$ subsets of $\Omega$ are produced
by removing a $U_s$ from $\Omega$, i.e. $\Omega_s=\Omega\setminus
U_s$. Finally, the MI estimator is applied on each $\Omega_s$ for the chosen
variables and a range of values to explore for $K$. For a fixed value of $K$
the $S$ obtained values $\widehat{MI}_s(X,Y)$ ($1\leq s\leq S$) provide a way
to estimate the variance of the estimator.

The bias problem is addressed in a similar way by providing some reference
value for the MI. Indeed if $X$ and $Y$ are independent variables, then
$MI(X,Y)=0$. Because of the bias (and variance) of the estimator, the
estimated value $\widehat{MI}(X,Y)$ has no reason to be equal to zero (it can
even be negative, whereas the mutual information is theoretically bounded
below by 0). However if some variables $X$ and $Y$ are known to be
independent, then the mean of $\widehat{MI}(X,Y)$ evaluated via a
cross-validation approach provides an estimate of the bias of the
estimator. In practice, given two variables $X$ and $Y$ known through
observations $\Omega =\{(x_n,y_n)\}, 1\leq n\leq N$, independence is obtained
by randomly permuting the $y_n$ without changing the $x_n$. Combined with the
cross-validation strategy proposed above, this technique leads to an
estimation of the bias of the estimator. Of course, there is no particular
reason for the bias to be uniform: it might depend on the actual value of
$MI(X,Y)$. However, a reference value if needed to obtain an estimate and the
independent case is the only one for which the true value of the mutual
information is known. The same $X$ and $Y$ as those used to calculate $\widehat{MI}_s(X,Y)$ should be of course be used, in order to remove from the bias estimation a possible dependence on the distributions (or entropies) of $X$ and $Y$; just permuting the same variables helps reducing the differences between the dependent case
and the reference independent one. 

The cross-validation method coupled with permutation provides two (empirical)
distributions respectively for $\widehat{MI}_K(X,Y)$ 
and $\widehat{MI}_K(X,\pi(Y))$, where $\pi$ denotes the permutation operation
and where the $K$ subscript is used to emphasises the dependency on $K$. A
good choice of $K$ then corresponds to a situation where the (empirical) variances
of $\widehat{MI}_K(X,Y)$  and $\widehat{MI}_K(X,\pi(Y))$ and the (empirical)
mean of $\widehat{MI}_K(X,\pi(Y))$ are small. Another way to formulate a
similar constraint is to ask for $\widehat{MI}_K(X,Y)$ to be significantly
different from $\widehat{MI}_K(X,\pi(Y))$ when $X$ and $Y$ are known to
exhibit some dependency. The differences between the two distributions can be
measured for instance via a measure inspired from Student's t-test. Let us
denote $\mu_K$ (resp. $\mu_{K,\pi}$) the empirical mean of $\widehat{MI}_K(X,Y)$
(resp. $\widehat{MI}_K(X,\pi(Y))$) and $\sigma_K$ (resp. $\sigma_{K,\pi}$) its
empirical standard deviation. Then the quantity
\begin{equation}
\label{eq:ttest}
t_K=\frac{\mu_K -\mu_{K,\pi}}{\sqrt{\sigma_K^2+\sigma^2_{K,\pi}}}
\end{equation}
measures the significance of the differences between the two (empirical)
distributions (if the distributions were Gaussian, a Student's t-test of
difference in the means of the distributions could be conducted). Then one
chooses the value of $K$ that maximizes the differences between the
distributions, i.e. the one that maximizes $t_K$.

The pseudo-code for the choice of $K$ in the MI estimator is given in Table
\ref{tab:ChoiceOfK}. In practice, the algorithm will be applied to each real
valued variable that constitute the $X$ vector and the
optimization of $K$ will be done along all the obtained $t_K$ values. As $t_K$
will be larger for relevant variables than for non-relevant ones, this allows to
discard automatically the influence of non-relevant variables in the choice of
$K$. 

\begin{table}[htbp]
  \centering
\texttt{
  \begin{tabbing}
 Inputs \=  $\Omega =\{(x_n,y_n)\}, 1\leq n\leq N$ the dataset \\
        \>  $K_{\min}$ and $K_{\max}$ the range where to look for the optimal
        $K$ \\
        \> $S$ the cross-validation parameter\\
Output  \> the optimal value of $K$ \\
\\\pushtabs
Code \> Dr\=aw a random partition of $\Omega$ into $S$ subsets
$U_1,\ldots,U_S$ \\   
     \> \> with roughly equal sizes\\
     \> Draw a random permutation $\pi$ of $\{1,\ldots,N\}$\\
\poptabs
     \> For \= $K\in \{K_{\min},\ldots,K_{\max}\}$\\
     \> \> For \= $s\in \{1,\ldots,S \}$\\\pushtabs
     \> \>      \> co\=mpute $mi[s]$ the estimation of the mutual information \\
     \> \>      \> \> $MI(X,Y)$ based on $\Omega_s=\Omega\setminus U_s$\\
     \> \>      \> compute $mi_\pi[s]$ the estimation of the mutual information \\
     \> \>      \> \> $MI(X,\pi(Y))$ based on $\Omega_s=\Omega\setminus U_s$\\
     \> \>      \> \> with the permutation $\pi$ applied to the $y_i$\\
     \> \> EndFor\\
\poptabs
     \> \> Compute $\mu_K$ the mean of $mi[s]$ and $\sigma_K$ its standard deviation \\
     \> \> Compute $\mu_{K,\pi}$ the mean of $mi_\pi[s]$ and $\sigma_{K,\pi}$
     its standard deviation \\
     \> \> Compute $t_K=\frac{\mu_K
       -\mu_{K,\pi}}{\sqrt{\sigma_K^2+\sigma^2_{K,\pi}}}$\\
     \> EndFor \\
     \> Return the smallest $K$ that minimises $t_K$ on $\{K_{\min},\ldots,K_{\max}\}$
  \end{tabbing}
}
  \caption{Pseudo-code for the choice of $K$ in the mutual information estimator}
  \label{tab:ChoiceOfK}
\end{table}

\begin{figure}[htbp]
  \centering
\includegraphics[angle=270,width=\textwidth]{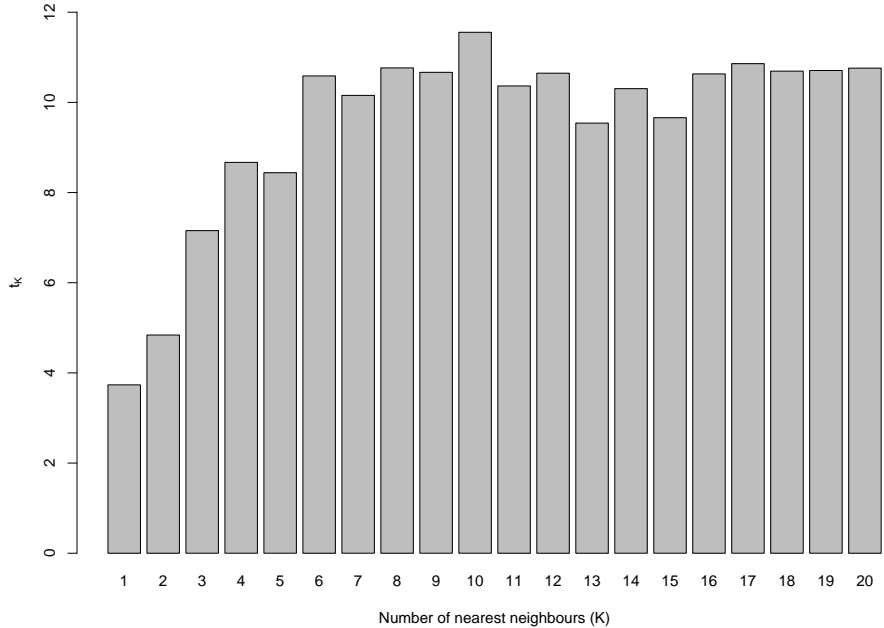}  
  \caption{Values of $t_K$ for $MI(X_4,Y)$ (see text for details)}
  \label{fig:friedman:tkv4}
\end{figure}

\begin{figure}[htbp]
  \centering
\includegraphics[angle=270,width=\textwidth]{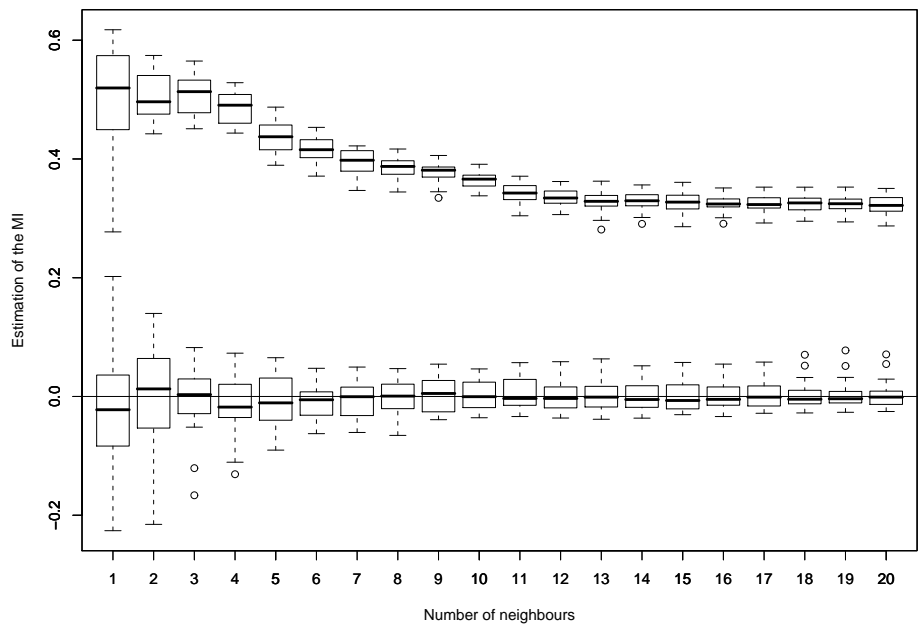}  
  \caption{Boxplots for the distributions of $\widehat{MI}_K(X_4,Y)$ (top) and
    of $\widehat{MI}_K(X_4,\pi(Y))$ (bottom) as a function of $K$}
  \label{fig:friedman:v4:simul}
\end{figure}

To test the proposed methodology, a dataset is generated as follows.  Ten
features $X_i, 1\leq i\leq 10$ are generated from a uniform distribution in
$[0,1]$.  Then, $Y$ is built according to 
\begin{equation}
\label{eq:example}
Y=10\sin(X_1 X_2)+20(X_3-0.5)^2+10X_4+5X_5+\epsilon,
\end{equation}
where $\epsilon$ is a Gaussian noise with zero mean and unit variance.  Note
that variables $X_6$ to $X_{10}$ do not enter into equation \eqref{eq:example};
they are independent from the output $Y$.  A sample size of 100 observations
is used and the CV parameter is $S=20$.  When evaluating the MI between $Y$
and a relevant feature (for 
example $X_4$), a $t_K$ value is obtained for each value of $K$, as shown on
Figure \ref{fig:friedman:tkv4}. Those values summarize the differences
between the empirical distributions of  $\widehat{MI}_K(X_4,Y)$  and
    of $\widehat{MI}_K(X_4,\pi(Y))$ (an illustration of the behaviour of those
    distributions is given in Figure \ref{fig:friedman:v4:simul}). The largest
$t_K$ value corresponds to the smoothing parameter $K$ that best separates the
distributions in the relevant and permuted cases (in this example the optimal
$K$ is 10).  

\subsection{Stopping criterion}
As mentioned above, stopping the greedy forward or forward-backward procedure
when the estimated MI decreases is not sound or theoretically justified.  A
better idea is to measure whether the addition of a feature to the already
selected set increases significantly the MI, compared to a situation where a
non-relevant feature is added, again in the same settings i.e. keeping the same
distribution for the potentially relevant variable and the non-relevant one.

This problem is similar to the previous one. Given a subset of already
selected variables $S$, a new variable $X_{st}$ is considered significant if
the value of $MI(S\cup X_{st},Y)$ significantly differs from the values
generated by $MI(S\cup \pi(X_{st}),Y)$, where $\pi$ is a random
permutation. In practice, one generates several random permutation and counts
the number of times that $MI(S\cup \pi(X_{st}),Y)$ is higher than $MI(S\cup
X_{st},Y)$. This gives an estimate of the p-value of $MI(S\cup X_{st},Y)$
under the null hypothesis that $X_{st}$ is independent from $(S,Y)$. A small
value means that the null hypothesis should be rejected and therefore that
$X_{st}$ brings significant new information about $Y$. The pseudo-code for the
proposed algorithm is given in Table \ref{tab:StoppingCriterion}. 

It should be noted that the single estimation of $MI(S\cup X_{st},Y)$ could be
replaced by a cross-validation based estimate of the distribution of this
value. The same technique should then be used in the estimation of the
distribution of $MI(S\cup \pi(X_{st}),Y)$. 

\begin{table}[htbp]
  \centering
\texttt{
  \begin{tabbing}
 Inputs \=  $\Omega =\{(x_n,y_n)\}, 1\leq n\leq N$ the dataset \\
        \> $P$ the number of permutations to compute\\
        \> the subset $S$ of currently selected variables\\
        \> the candidate variable $X_{st}$\\
Output  \> a p-value for the hypothesis that the variable is useless \\
Code \> Compute $ref$ the value of $MI(S\cup X_{st},Y)$\\
     \> Initialise $out$ to 0\\
     \> For \= $p\in \{1,\ldots,P\}$\\
     \>     \>  Draw a random permutation $\pi_p$ of $\{1,\ldots,N\}$\\
     \>     \> If \= $MI(S\cup \pi_p(X_{st}),Y)\geq ref$ then \\
     \>     \>    \> increase $out$ by $1$\\
     \>     \> EndIf\\
     \> EndFor\\
     \> Return $out/P$
  \end{tabbing}
}
  \caption{Pseudo-code for the choice of the stopping criterion for the greedy procedure}
  \label{tab:StoppingCriterion}
\end{table}

To illustrate this method, 100 datasets are randomly generated according to
equation \eqref{eq:example}. For each dataset, the optimal value of $K$ for
the MI estimator is selected according to the method proposed in Section
\ref{subsection:Kchoice}. Then a forward procedure is applied and stopped
according to the method summarized in Table \ref{tab:StoppingCriterion} (with
a significance threshold of $0.05$ for the p-value). As it can be seen from
Table \ref{tab:NumberOfSelectedFeatures}, in most cases 4 or 5 relevant
features are selected by the procedure (5 is the expected number, as $X_6$ to
$X_{10}$ are not linked with $Y$).  Without resampling, by stopping the
forward procedure at the maximum of mutual information, in most cases only 2
(45 cases) and 3 (33 cases) features are selected.  This is a consequence of
the fact that when looking only to the value of the estimated MI at each step,
the estimation is made in spaces of increasing dimension (the dimension of $X$
is incremented at each step).  It appears that in average the estimated MI
decreases with the dimension, making irrelevant the comparison of MI estimations with
feature vectors of different dimensions.

\begin{table}
\centering
\begin{tabular}{lcccccc}
  \hline
  Number of features & 1 & 2 & 3 & 4 & 5 & 6 \\
  Percentage & 0 & 1 & 12 & 52 & 29 & 6 \\
  \hline
\end{tabular}
	\caption{Number of selected features}
	\label{tab:NumberOfSelectedFeatures}
\end{table}

More experiments on the use of resampling to select $K$ and to stop the
forward procedure may be found in
\cite{neurocomp}. 

\subsection{Clustering by rank correlation}

In some problems and applications, the number of features that are relevant for the prediction of variable $Y$ may be too large to afford the above described procedures.  Indeed, as detailed in  the previous sections, the estimator of mutual information will fail when used on too high-dimensional variables, despite all precautions that are taken (using an efficient estimator, avoiding a backward procedure, using estimator results on a comparative basis rather than using the rough values, etc.).

In this case, another promising direction is to cluster features instead of selecting them \cite{cluster,VanDijk:2006}.  Feature clustering consists in grouping features in natural clusters, according to a similarity measure: features that are similar should be grouped in a single cluster, in order to elect a single representative from the latter.  This is nothing else than applying to features the traditional notion of clustering usually applied to objects.  For example, all hierarchical clustering methods can be used, the only specific requirement being to define a measure of similarity between features.  Once the measure of similarity is defined, the clustering consists in selecting the two most similar features and replacing them by a representative.  Next, the procedure is repeated on the remaining initial features and representatives.

The advantage of feature clustering with respect to the procedure based on the mutual information between a group of features and the output, as described above, is that the similarity measure is only used on two features (or representatives) at each iteration.  Therefore the problems related to the increasing dimensionality of the feature sets is completely avoided.  The reason behind this advantage is that in the first case the similarity is measured between a set of features and the output, while in the clustering the similarity is measured between features (or their representatives) only.  Obviously, the drawback is that the variable $Y$ to predict is no more taken into account.

In order to remedy to this last problem, a new conditional measure of similarity between features is introduced.  Simple correlation or mutual information between features could be used, but will not take the information from $Y$ into account.  However, based on the idea of Kraskov's estimator of Mutual Information, one can define a similarity measure that takes $Y$ into account, as follows \cite{agrostat}.

\begin{figure}[htbp]
\label{fig:rank_neigbors}
  \centering
\includegraphics[width=\textwidth]{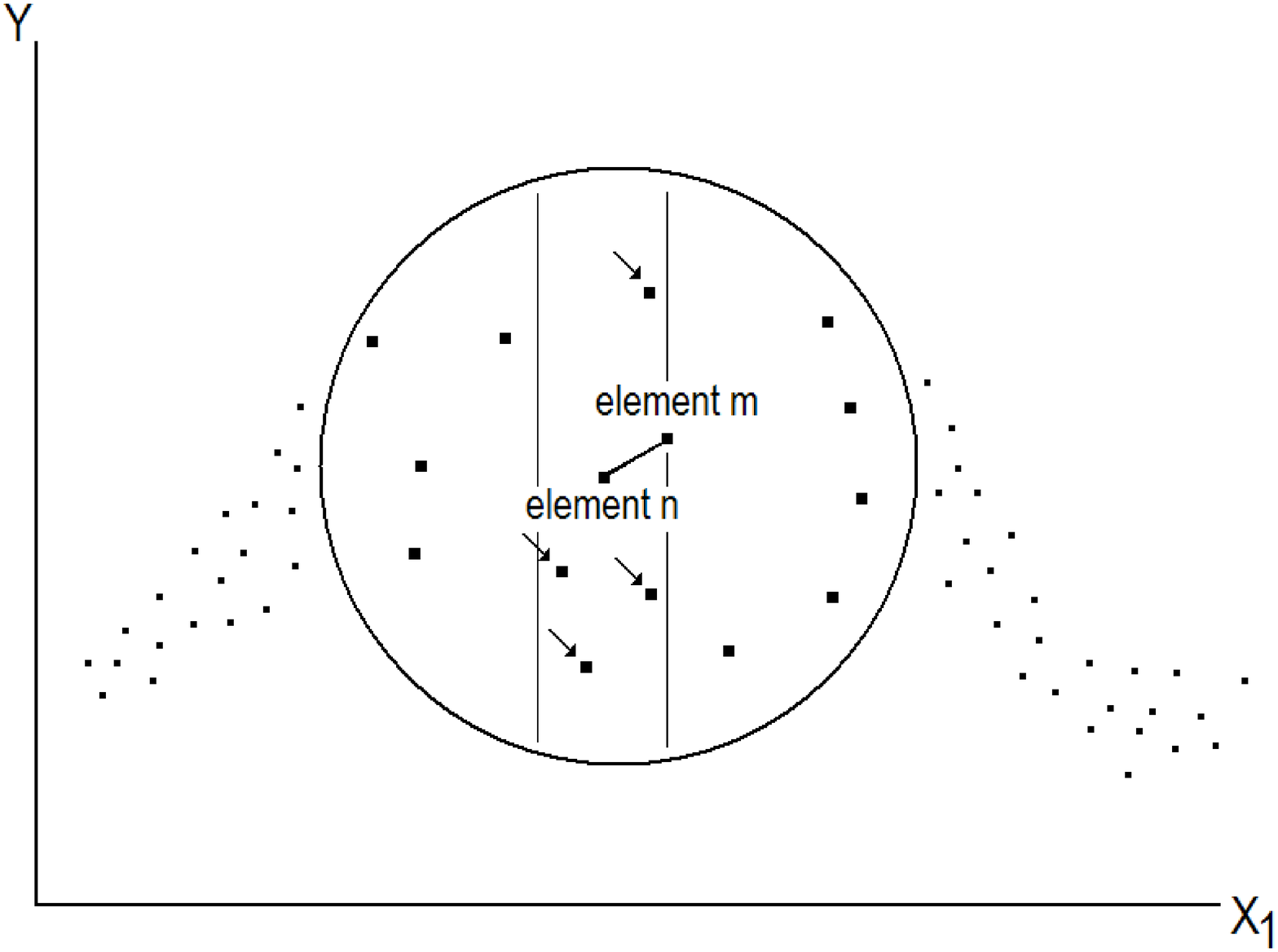}  
\caption{Identification of neighbors in $X_1$ that are not neighbors in $(X_1,Y)$.}
\end{figure}

Let $X_1$ and $X_2$ the two features whose similarity should be measured.  The idea is to measure if $X_1$ and $X_2$ contribute similarly, or not, to the prediction of $Y$.  Let $\Omega={z_n=(x_{1n},x_{2n},y_n)}, 1\leq n\leq N$ the sample set; other features than $X_1$ and $X_2$ are discarded from the notation for simplicity.  For each element $n$, we search for the nearest neighbor according to the Euclidean distance in the joint $(X_1,Y)$ space.  Let we denote this element by its index $m$.  Then, we count the number $c1_n$ of elements that are closer from element $n$ than element $m$, taking only into account the distance in the $X_1$ space.  Figure \ref{fig:rank_neigbors} shows such elements.  $c1_n$ is a measure of the number of local false neighbors, i.e. elements that are neighbors according to $X_1$ but not according to $(X_1,Y)$.  If this number if high, it means that element $n$ can be considered as a local outlier in the relation between $X_1$ and $Y$.

The process is repeated for all elements $n$ in the sample set, and resulting $c1_n$ values are concatenated in a $N$-dimensional vector $C1$.  Next, the same procedure is applied with feature $X_2$ instead of $X_1$; resulting $c2_n$ values are concatenated in vector $C2$.

If features $X_1$ and $X_2$ carry the same information to predict $Y$, vectors $C1$ and $C2$ will be similar.  On the contrary, if they carry different yet complementary information, vectors $C1$ and $C2$ will be quite different.  Complementary information can be for example that $X_1$ is useful to predict $Y$ in a part of its range, while $X_2$ plays a similar role in another part of the range.  As the $c1_n$ and $c2_n$ vectors contain local information in the $(X_1,Y)$ (respectively $(X_2,Y)$) relation, vectors $C1$ and $C2$ will be quite different in this case.  For these reasons, the correlation between $C1$ and $C2$ is a good indicator of the similarity between $X_1$ and $X_2$ when these features are used to predict $Y$.  This is the similarity measure that is used in the hierarchical feature clustering algorithm.

In order to illustrate this approach, it is applied to two feature clustering problems where the number of initial features is large.  Analysis of (infrared) spectra is a typical example of such problem.  The first dataset, Wine cite{wine}, consists in 124 near-infrared spectra of wine samples, for which the concentration in alcohol has to be predicted from the spectra.  Three outliers are removed, and 30 spectra are kept aside for test.  The second dataset is the standard Tecator benchmark \cite{Borggaard:1992}; it consists of 215 near-infrared spectra of meat samples, 150 of them being used for learning and 65 for test.  The prediction model used for the experiments is Partial Least Squares Regression (PLSR); the number of components in the PLSR model is set by 4-fold cross-validation on the training set.  Three experiments are conducted on each set: a PLSR model on all features, a PLSR model on traditional clusters built without using $Y$, and a PLSR model built on clusters defined as above.  The results are shown in Table \ref{tab:ClusteringResults}; the Normalized Mean Square Error (NMSE) on the test set is given, together with the number of features or clusters.

\begin{table}
\label{tab:ClusteringResults}
\caption{Results of the feature clustering on two spectra datasets}
\begin{tabular}{cccc}
 & without clustering & clustering without $Y$ & clustering with $Y$ \\
\hline
Wine & $NMSE=0.00578$ & $NMSE=0.0111$ & $NMSE=0.00546$ \\
 & 256 features & 19 clusters & 33 clusters \\
\hline
Tecator & $NMSE=0.02658$ & $NMSE=0.02574$ & $NMSE=0.02550$ \\
 & 100 features & 17 clusters & 8 clusters \\
\end{tabular}
\end{table}

In both cases, the clustering using the proposed method (last column) performs better than a classical feature clustering, or no clustering at all.  In the Tecator experiment, the advantage in terms of performances with respect to the non-supervised clustering is not significant; however, in this case, the number of resulting clusters is much smaller in the supervised case, which reaches the fundamental goal of feature selection, i.e. the ability to build simple, interpretable models.

\section{Conclusion}

Feature selection in supervised regression problems is a fundamental preprocessing step.  Feature selection has two goals.  First, similarly to other dimension reduction techniques, it is aimed to reduce the dimensionality of the problem without significant loss of information, therefore acting against the curse of dimensionality.  Secondly, contrarily to other approaches where new variables are built from the original features, feature selection helps to interpret the resulting prediction model, by providing a relevance measure associated to each original feature.

Mutual Information (MI), a concept borrowed from information theory, can be used for feature selection.  The MI criterion is used to test the relevance of subsets of features with respect to the prediction task, in a greedy procedure.  However, in practise, the MI theoretical concept needs to be estimated.  Even if efficient estimators exist, they still suffer from two drawbacks: their performances decrease with the dimension of the feature subsets, and they need to set a smoothing parameter (for example $K$ in a $K$-nearest neighbors based estimator).

In addition, when embedded in a forward selection procedure, the MI does not provide any stopping criterion, at least in theory.  Standard practice to stop the selection when the estimation of the MI begins to decrease exploits in fact a limitation of the estimator itself, without any guarantee that the algorithm will indeed stop when no further feature has to be added. 
 
This chapter shows how to cope with these three limitations.  It shows how using resampling and permutations provides first a way to compare MI values on a sound basis, and secondly a stopping criterion in the forward selection process.

In addition, when the number of relevant features is high, there is a need to avoid using MI between feature sets and the output, because of the too high dimension of the feature sets.  It is also shown how to cluster features by a similarity criterion used on single features.  The proposed criterion measures whether two features contribute identically or in a complementary way to the prediction of $Y$; the measure is thus supervised by the prediction task.

These methodological proposals are shown to improve the results of a feature selection process using similarity measures based on Mutual Information.

\bibliographystyle{abbrv}
\bibliography{paperMV}

\end{document}